\pdfoutput=1

\documentclass[11pt]{article}

\usepackage[]{ACL2023}

\usepackage{times}
\usepackage{latexsym}

\usepackage[T1]{fontenc}

\usepackage[utf8]{inputenc}

\usepackage{microtype}

\usepackage{inconsolata}

\usepackage{booktabs} 
\usepackage{graphicx}  
\usepackage{multirow}
\usepackage{subfigure}
\usepackage{xspace}
\usepackage{colortbl}
\usepackage{xcolor}
\usepackage{pifont}
\usepackage{amsmath}
\usepackage{amsfonts}
\usepackage{resizegather}
\usepackage{algorithm}
\usepackage{algorithmic} 
\usepackage{amssymb}
\usepackage{svg}

\newcommand{\aka}{\emph{a.k.a.,}\xspace}
\newcommand{\eg}{\emph{e.g.,}\xspace}

\newcommand{\ignore}[1]{}

\newcommand{\dubbelop}{$^{\blacktriangle}$}

\newcommand{\dubbelneer}{$^{\blacktriangledown}$}

\newcommand{\fullmodel}{\textbf{U}nified \textbf{M}ulti-scenario \textbf{S}ummarization \textbf{E}valuation Model\xspace}
\newcommand{\model}{UMSE\xspace}

\usepackage[hang]{footmisc}
\title{\model: Unified Multi-scenario Summarization Evaluation}

\author{Shen Gao$^1$\thanks{~Equal contribution.}~~~Zhitao Yao$^{1*}$~~~Chongyang Tao$^2$~~~Xiuying Chen$^3$~~~Pengjie Ren$^1$\\\textbf{Zhaochun Ren}$^1$~~~\textbf{Zhumin Chen}$^1$\thanks{~Corresponding author.} \\
        $^1$Shandong University, Qingdao, China\\
        $^2$Microsoft Corporation, Beijing, China\\
        $^3$King Abdullah University of Science and Technology, Thuwal, Saudi Arabia\\
        \{shengao,renpengjie,zhaochun.ren,chenzhumin\}@sdu.edu.cn, yaozhitao@mail.sdu.edu.cn\\
         chotao@microsoft.com, xiuying.chen@kaust.edu.sa
        }

\begin{document}
\maketitle
\begin{abstract}
Summarization quality evaluation is a non-trivial task in text summarization.
Contemporary methods can be mainly categorized into two scenarios: (1) \textit{reference-based}: evaluating with human-labeled reference summary; (2) \textit{reference-free}: evaluating the summary consistency of the document.
Recent studies mainly focus on one of these scenarios and 
explore training neural models built on pre-trained language models (PLMs) to align with human criteria.
However, the models from different scenarios are optimized individually, which may result in sub-optimal performance since they neglect the shared knowledge across different scenarios. Besides, designing individual models for each scenario caused inconvenience to the user.
Inspired by this, we propose \fullmodel (\model).
More specifically, we propose a perturbed prefix tuning method to share cross-scenario knowledge between scenarios and use a self-supervised training paradigm to optimize the model without extra human labeling.  
Our \model is the first unified summarization evaluation framework engaged with the ability to be used in three evaluation scenarios.
Experimental results across three typical scenarios on the benchmark dataset SummEval indicate that our \model can achieve comparable performance with several existing strong methods which are specifically designed for each scenario.\footnote{~Code is available at https://github.com/ZT-Yao/UMSE.}

\end{abstract}

\section{Introduction}

Quantitatively evaluating the quality of generated summary is a non-trivial task that can measure the performance of the summarization system~\cite{Lin2004Rouge,NG2015Rougewe,Zhang2020BERTScore,Thomas2021QuestEval}, and can also be used as a reward model to give an additional training signal for the summarization model~\cite{Wu2018RLforsum,Sha2018RLforsum,Thomas2019SummaQA,Gao2019RASG,Gao2020Summarization}.
The dominant evaluation methods are traditional word-overlap-based metrics like ROUGE~\cite{Lin2004Rouge} and BLEU~\cite{Papineni2002Bleu}.
Although these metrics are very easy to use, they cannot evaluate semantic similarity.
In recent years, many researchers focus on semantic-based evaluation tools~\cite{NG2015Rougewe,Zhang2020BERTScore,Zhao2019MoverScore}.
Different to traditional metrics which only use one score to measure the quality of the summary, \citet{Zhong2022UniEval} propose to evaluate the summary quality in several dimensions (\eg coherence, consistency, and fluency) by calculating the similarity between the generated summary and the human-annotated summary.

\begin{figure}
\centering
\includegraphics[width=0.8\columnwidth]{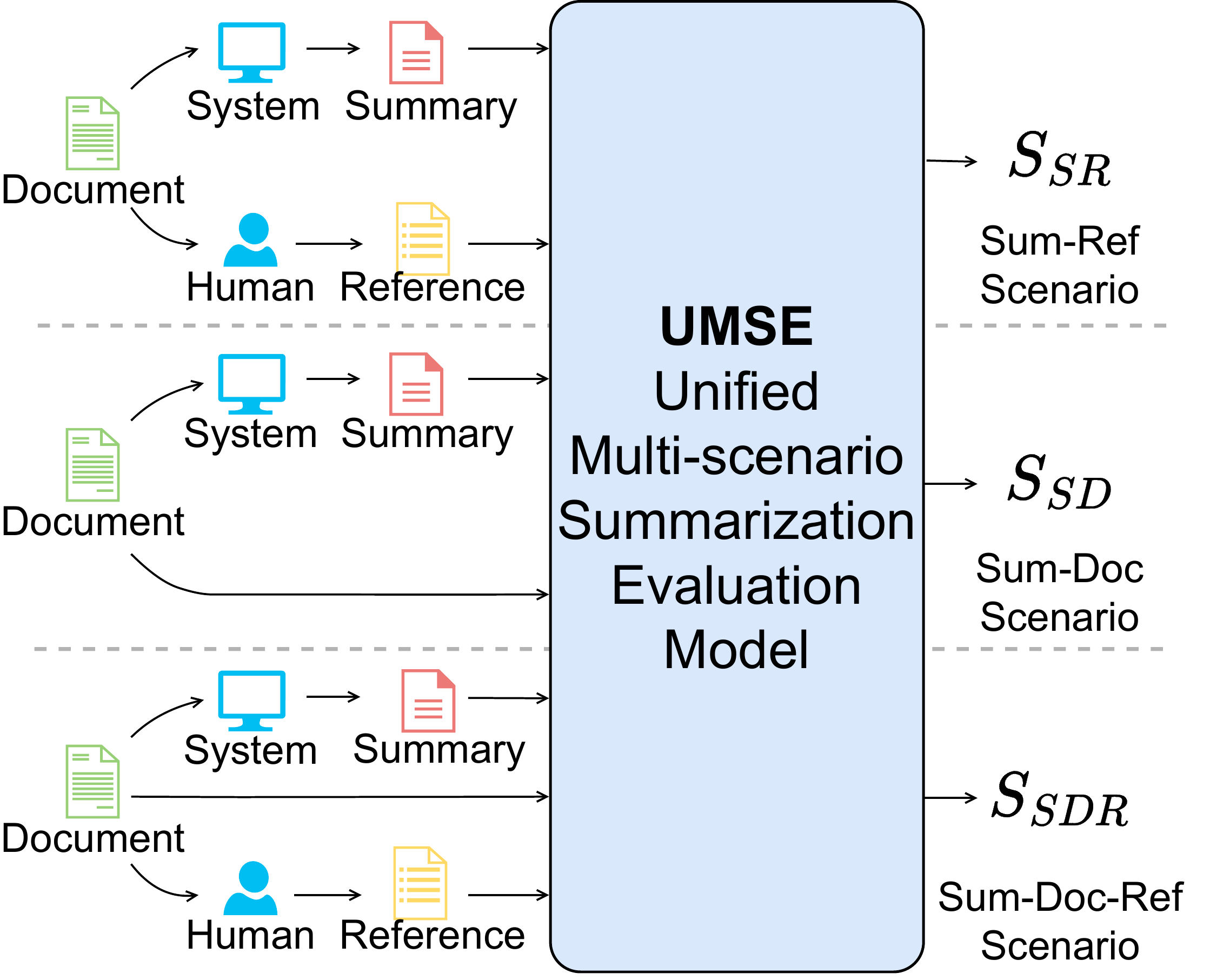}
\caption{Illustration of multi-scenario summarization evaluation.}
\label{fig:model-intro}
\end{figure}


The summarization evaluation methods can be categorized into two scenarios based on the input data type: (1) \textbf{reference-based} methods require the human-annotated summary as input and (2) \textbf{reference-free} methods only use the corresponding document.
The reference-based methods~\cite{Lin2004Rouge,Papineni2002Bleu,Banerjee2005METEOR,NG2015Rougewe,Zhang2020BERTScore,Zhao2019MoverScore,Yuan2021BARTScore} usually use the human written summary (\aka reference summary) as the ground truth and calculate the similarity between generated and reference summary.
With the help of the pre-train language model, these methods have a powerful ability to measure semantic similarity.
However, not all real-world application scenarios have human-annotated summaries.
Using the reference-based evaluation method with the human-annotated ground truth summary is labor-consuming.
Thus, reference-free methods~\cite{Wu2020LS_Score,Gao2020SUPERT,Thomas2019SummaQA,Thomas2021QuestEval} propose to evaluate the summary by modeling the semantic consistency between the generated summary and the document.

When evaluating a summarization system, even though we can individually select a proper evaluator condition on whether we have a reference summary, it is not very convenient.
Moreover, since human annotation is costly, some summarization methods~\cite{Wu2018RLforsum,Sha2018RLforsum,Thomas2019SummaQA} choose to use the automatic evaluator to provide an additional training signal, instead of relying entirely on human-labeled document-summary pair data.
In this type of usage, the evaluator needs to measure the quality of the model-generated summary with \textit{partial} human-labeled document-summary data.
Besides, contemporary trainable evaluation models for different scenarios (with or without reference summary) are built on pre-train language models, which may transfer knowledge across different scenarios and provides a great opportunity to bridge these evaluation scenarios with a better combination of the best of both worlds.
Hence, it is valuable to build a unified multi-scenario summarization evaluator that can be used for processing both types of input data.
Intuitively, this naturally leads to two questions: (1) \textit{How to build a unified multi-scenario evaluation model regardless of whether we have a reference summary?} (2) \textit{How to train the evaluator so that it can share knowledge between scenarios and maintain the exclusive knowledge in a specific task?}

In this paper, we propose a unified multi-scenario summarization evaluation method \fullmodel (\model).
\model unifies three typical summary quality evaluation scenarios in one model: 
(1) \textbf{Sum-Ref}: evaluate using reference summary. \model measures the similarity between the generated summary and the human-annotated reference summary.
(2) \textbf{Sum-Doc}: evaluate using document. Since using the reference summary is labor-consuming, \model can measure the consistency between generated summary and the original document.
(3) \textbf{Sum-Doc-Ref}: evaluate using both document and reference summary. This method incorporates the advantages of sum-ref and sum-doc.
To process these different types of input, we propose a perturbed prefix method based on the prefix tuning method~\cite{Liang2021Prefix,Liu2022Ptuningv1,Liu2021Ptuningv2} that shares a unified pre-train language model across three scenarios by using different continuous prefix tokens as input to identify the scenario.
Then, we propose $2$ hard negative sampling strategies to construct a self-supervised dataset to train the \model without additional human annotation.
Finally, we propose an ensemble paradigm to combine these scenarios into a unified user interface.

To sum up, our \model can bring the following benefits:

$\bullet$ \textbf{One model adaptable to multi-scenario}. \model uses only one model to evaluate the generated summary whenever it has a reference summary.

$\bullet$ \textbf{Mutually enhanced training}. We propose a perturbed prefix method to transfer knowledge between scenarios, and it can boost the performance of each scenario.

$\bullet$ \textbf{Self-supervised}. \model can be trained using a fully self-supervised paradigm without requiring any human-labeled data, and it makes \model has strong generalization ability.

To verify the effectiveness of the \model, we first compare with several baselines including the reference-based and reference-free methods.
Specifically, \model outperforms all the strong reference-free evaluation methods by a large margin and achieves comparable performance with the state-of-the-art in a unified model.
Ablation studies verify the effectiveness of our proposed perturbed prefix-tuning method.

\section{Related Work}\label{sec:related}

\subsection{Reference-free Metrics} 

Reference-free metrics aim to evaluate the summary quality without the human-labeled ground truth summary as the reference, and these methods can be categorized into two types: trained model and training-free model. 
For the training-free methods, SUPERT~\cite{Gao2020SUPERT} first extracts salient sentences from the source document to construct the pseudo reference, then computes the semantic similarity to get the evaluation score. 
Following SUPERT, \citet{Chen2021Multi_Doc_Metric} propose a centrality-weighted relevance score and a self-referenced redundancy score. 
While computing the relevance score, the sentences of pseudo reference are weighted by centrality, the importance of each sentence.
For the methods which should be trained, LS-Score~\cite{Wu2020LS_Score} is an unsupervised contrastive learning framework consisting of a linguistic quality and a semantic informativeness evaluator. 
The question-answering paradigm is usually used in evaluating summaries, which evaluates the factual consistency between summary and document with the help of well-trained question-answering models~\cite{Thomas2019SummaQA,Gao2019PAAG,Esin2020FEQA,Thomas2021QuestEval}. 

\subsection{Reference-based Metrics}

Referenced-based metrics, which evaluate the quality of the summary by measuring the similarity of the summary and human written reference, can be divided into two categories: lexical overlap-based metrics and semantic-based metrics.
ROUGE~\cite{Lin2004Rouge}, the most commonly used metric for summary evaluation, measures the number of matching n-grams between the system output and reference summary. 
Other popular lexical overlap-based metrics are BLEU~\cite{Papineni2002Bleu} and METEOR~\cite{Banerjee2005METEOR} which are also commonly employed in other text generation tasks (\eg machine translation). 
Since using the lexical overlap to measure the quality is sometimes too strict, many researchers turn to focus on exploring the semantic-based evaluation.
ROUGE-WE~\cite{NG2015Rougewe} improves ROUGE by using Word2Vec~\cite{Tomas2013Word2vec} embeddings, and S3~\cite{Maxime2017S3} takes the ROUGE and ROUGE-WE as input features and is trained on human-annotated datasets. 
With the prosperity of the pre-training language model (PLM), more and more researchers introduce these models for evaluation. 
BERTScore~\cite{Zhang2020BERTScore} leverages the contextual embeddings from BERT~\cite{Devlin2019BERT} and calculates the cosine similarity between system output and reference sentence. 
CTC~\cite{Degn2021CTC} is based on information alignment from two dimensions: consistency and relevance. 
UniEval~\cite{Zhong2022UniEval} is a multi-dimensional evaluator based on T5~\cite{2020t5}, and it formulates the summary evaluation as a binary question-answering task and evaluates from four dimensions: coherence, consistency, fluency, and relevance.
However, existing summarization evaluation models usually focus on measuring the summary quality from multiple aspects and transferring knowledge from PLM, they ignore the shareable knowledge between different scenarios.

Evaluating the quality of the generated text is a also crucial task in generation tasks.
In machine translation evaluation, \citet{Yu2022UniTE} proposes UniTE which is a multi-scenario evaluation method.
UniTE employs monotonic regional attention to conduct cross-lingual semantic matching and proposes a translation-oriented synthetic training data construction method.
However, the summarization task does not have these characteristics and directly applying UniTE to summarization evaluation cannot measure the important aspect of summary (\eg coherence and relevance).

\definecolor{modelcolor}{RGB}{176, 115, 158}
\definecolor{datacolor}{RGB}{226, 106, 100}
\definecolor{scenacolor}{RGB}{132, 167, 226}
\begin{figure*}[htb]
\centering
\includegraphics[width=2.0\columnwidth]{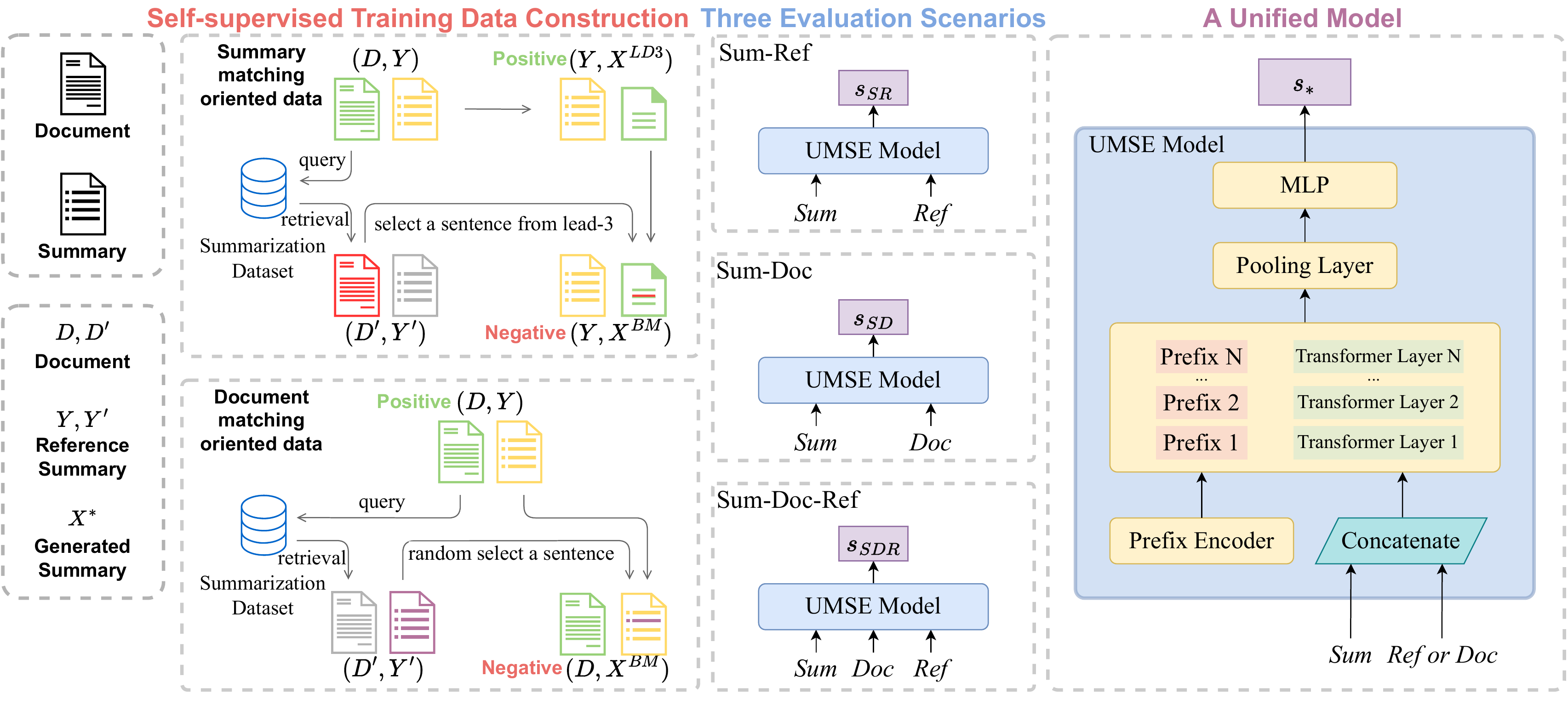}
\caption{Illustration of \model which tackles the summarization evaluation in three scenarios by a unified model trained with two self-supervised tasks.}
\label{fig:model}
\end{figure*}

\section{\model Model}

\paragraph{Problem Formulation}
Given a model-generated summary $X = \{x_1, x_2, \dots, x_{L_x}\}$ with $L_x$ tokens, our goal is to use a unified evaluation model to produce a score $s \in \mathcal{R}$ for $X$.
For the Sum-Ref scenario, the model uses generated summary $X$ and ground truth summary $Y = \{y_1, y_2, \dots, y_{L_y}\}$ as input.
For the Sum-Doc scenario, we evaluate the summary quality by using generated summary $X$ and document $D = \{d_1, d_2, \dots, d_{L_d}\}$ with $L_d$ tokens as input, which does not require any human annotation (\eg ground truth summary $Y$).
For the Sum-Doc-Ref scenario, the model uses generated summary $X$, ground truth summary $Y$, and document $D$ as input.
To train the evaluation model, we do not use any human-annotated summary quality dataset and we construct the training dataset by using several self-supervised training strategies.

\subsection{Overview}

In this section, we detail the \fullmodel (\model).
An overview of \model is shown in Figure~\ref{fig:model}.
\model has two main parts:
(1) \textbf{Data construction.} We first construct two self-supervised datasets for coherence and relevance evaluation scenarios. 
(2) \textbf{Unified Model.} To unify the different input data into a unified model, we propose a perturbed prefix-tuning method to train the \model.


\subsection{Data Construction}

Employing a human annotator to annotate the quality of generated summary to train the evaluation model is labor-consuming and will lead the evaluation model hard to use.
We propose to use the self-supervised tasks to construct the training dataset for the evaluator without using any human annotation.
Since measuring the quality of the summary requires two main semantic matching abilities: (1) matching with the reference summary and (2) matching with the document,
we propose two self-supervised tasks to construct the training dataset automatically:

\noindent $\bullet$ \textbf{Summary matching oriented data}: The goal for this task is to construct positive and negative samples which are different in whether the summary contains the salient information.
Given a document-summary pair $D, Y$, the data sample to construct is a summary pair.
The positive data pair $(Y, X^{LD3})$ contains the reference summary $Y$ and a candidate summary $X^{LD3}$ which contains relevant information.
And the negative data pair $(Y, X^{BM})$ contains the reference summary $Y$ and a candidate summary $X^{BM}$ which describes similar but not relevant information.
Particularly, if the negative data is very hard for the evaluation model to identify (\eg requires reasoning ability or is very similar to the positive sample), the evaluation model will achieve better performance than using very simple negative data.
Thus, we propose to use the leading three sentences of the corresponding document $D$ as the candidate summary $X^{LD3}$.
For the candidate summary $X^{BM}$ in negative data pair, we first use the BM25 retrieval model to retrieve the most similar document $D^\prime$ to $D$ and obtain the reference summary $Y^\prime$ of $D^\prime$.
To make the negative sample harder, we randomly replace a sentence in $Y^\prime$ with one sentence in $X^{LD3}$ as the final negative summary $X^{BM}$.

\noindent $\bullet$ \textbf{Document matching oriented data}: The golden criterion for evaluating the summary quality is whether the summary describes the main facts of the document. 
Hence, we construct self-supervised data which aims to train the model to measure the semantic relevance between summary and document.
The positive data pair $(D, Y)$ consists of document $D$ and its reference summary $Y$.
The negative data pair $(D, X^{BM})$ contains the document $D$ and a false summary $X^{BM}$ which is similar to $Y$.
We employ the same BM25 retrieval method in coherence data construction to obtain $Y^\prime$ and replace a sentence in $Y$ with a sentence in $Y^\prime$ as the negative summary $X^{BM}$.

For brevity, we omit the superscript of $X$ in the following sections.


\subsection{Perturbed Prefix-Tuning}

Although the three scenarios have different input types, we can directly concatenate them into a text sequence which can be easily adopted by the pre-train language model.
Following previous work~\cite{Zhong2022UniEval}, although our evaluation model does not require additional summarization-quality data annotations, human-written summaries are still required to train the estimator. 
Therefore, reducing the dependence on human-written summaries can improve the applicability of our model in low-resource scenarios.
Thus, we employ prefix-tuning to explore the semantic understanding ability of large language models on the summarization evaluation task.
Specifically, we append different prefix sequences at the start of each input text sequence according to the scenario:
\begin{align*}
\mathbf{H}_{SR} &= \text{PLM}(\text{[CLS]} P_{SR} X \text{[SEP]} Y), \\
\mathbf{H}_{SD} &= \text{PLM}(\text{[CLS]} P_{SD} X \text{[SEP]} D), \\
\mathbf{H}_{SDR} &= \text{PLM}(\text{[CLS]} P_{SDR} X \text{[SEP]} D \text{[SEP]} Y),
\end{align*}
where $\text{[CLS]}$ and $\text{[SEP]}$ are both special tokens in PLM, $\mathbf{H}_{SR} \in \mathbb{R}^{(L_x+L_y+L_p+2), z}$ denotes the token level representation for Sum-Ref pair, and $z$ is the hidden size of the PLM.
The $\bf P_{*} \in \mathbb{R}^{L_p, z}$ denotes the prefix for each scenario, which is a continuous prompt with length $L_p$.
The advantage of using the unified evaluator is that we can use one large language model to conduct three tasks and it will reduce the size of the evaluation toolkit.

Although these data scenarios have their exclusive task characteristic, there are also some shared abilities and knowledge which can be transferred between different scenarios.
To model the exclusive characteristic and transfer knowledge using the continuous prefix in a coordinated way, we propose a prefix perturbation method that uses the same tokens with different orders of different scenarios.
Take the prefix of Sum-Doc scenario as an example, $\bf P_{SD}$ contains $L_p$ continuous prefix tokens $\bf P_{SD} = \{\bf p_1, \bf p_2, \dots, \bf p_{L_p}\}$.
We perturb $\bf P_{SD}$ as $\{\bf p_1, \bf p_3, \dots, \bf p_{L_p}, \bf p_2, \dots, \bf p_{L_p-1}\}$, and use this perturbed prefix as the prefix for Sum-Doc-Ref $\bf P_{SDR}$.
This prefix perturbation method keeps the prefix used across scenarios to use the same continuous tokens in a different order.
Thus, our model can simultaneously transfer knowledge between scenarios and keep the exclusive ability prompted by the different prefixes.

To obtain the summary-level overall representation, we conduct a pooling operation on the token-level representation:
\begin{align}
\mathbf{E}_{SR}&=\text{Pooling}(\mathbf{H}_{SR}), \\
\mathbf{E}_{SD}&=\text{Pooling}(\mathbf{H}_{SD}), \\
\mathbf{E}_{SDR}&=\text{Pooling}(\mathbf{H}_{SDR}),
\end{align}
where $\mathbb{E}_{*} \in \mathbb{R}^z$ denotes the summary-level representation.
Then we employ a multi-layer perception (MLP) network to conduct a binary classification and obtain the probability $p$:
\begin{align}
p_*&=\text{Softmax}(\text{MLP}(\mathbf{E}_{*})) \in \mathcal{R}^2, \\
s_*&= p_*^+,
\end{align}
where $p_*^+ \in \mathbb{R}$ denotes the probability of positive class in $p_*$.
During training, we use cross entropy loss $\mathcal{L}_{ce}$ to optimize the model parameters to distinguish the positive and negative samples:
\begin{equation*}
\begin{aligned}
    \mathcal{L}_{ce} &=-\left[\sum_{i=1}^n c_i \log  p_i^{+}+\left(1-c_i\right) \log \left(1-p_i^+ \right)\right],
\end{aligned}
\end{equation*}
where $c_i \in \{0,1\}$ denotes the label of $i$-th training sample which indicates whether this sample is a positive or negative sample.
At the inference stage, we take the probability of positive class $p^+$ as the final evaluation score $s$.

\subsection{Variant of Sum-Doc-Ref Evaluation}\label{sec:sdr-variant}

Intuitively, the scenario Sum-Doc-Ref can be seen as a combination of the Sum-Doc and Sum-Ref scenarios.
Hence, an intuitive method to conduct the evaluation of the Sum-Doc-Ref scenario is to directly fuse the scores of the Sum-Doc and Sum-Ref scenarios.
In this section, we propose a variant implementation to conduct evaluation conditions on the input of Sum-Doc-Ref, named \textbf{\model {\scriptsize (Fusion)}}.
We combine the score of the Sum-Doc and Sum-Ref scenarios to get the score for the Sum-Doc-Ref:
\begin{equation}
s_{SDR}=f(s_{SR}, s_{SD}),
\end{equation}
where $f$ denotes the ensemble strategy, such as min and max. 
In the experiment, we will analyze the performance of different implementations of $f$. 
\newcommand{\cbkgrnd}{\cellcolor{blue!15}}
\newcommand{\sbbkgrnd}{\cellcolor{gray!35}}
\newcommand{\phantomtriangle}{\phantom{\dubbelop}}

\section{Experiment}\label{sec:exp-setup}







\subsection{Datasets}

In the training phase, we construct the positive and negative data pairs using the CNN/DailyMail~\cite{Nallapati2016CNN/DM} dataset. 
Then the trained evaluators are tested on the meta-evaluation benchmark SummEval~\cite{Fabbri2021Summeval} to measure the rank correlation coefficient between the evaluation model and human judgment.

\textbf{CNN/DailyMail} has $286,817$ training document-summary pairs, $13,368$ validation and $11,487$ test pairs in total. 
The documents in the training set have $766$ words and $29.74$ sentences on average while the reference summaries contain $53$ words and $3.72$ sentences.

\textbf{SummEval} is a meta-evaluation benchmark. To collect the human judgments towards the model-generated summaries, they first randomly select 100 document and reference pairs from the test set of CNN/DailyMail, then generate summaries using 16 neural summarization models. Each summary is annotated by 3 experts and 5 crowd-sourced workers along four dimensions: coherence, consistency, fluency, and relevance.
Finally, there is a total of $12800$ summary-level annotations.

\subsection{Evaluation Metrics}

Following previous work~\cite{Yuan2021BARTScore,Zhong2022UniEval}, we measure the rank correlation coefficient between the evaluation model and human judgment to represent the performance of the evaluator.
In the experiments, we employ the Spearman $(\rho)$ and Kendall-Tau $(\tau)$ correlations between the evaluator output scores and human ratings.
The statistical significance of differences observed between the performance of \model and the strongest baseline in each scenario is tested using a two-tailed paired t-test and is denoted using \dubbelop\ (or \dubbelneer) for strong significance at $\alpha=0.01$ and $p < 0.05$.

\subsection{Comparisons}
In the experiment, we compare the proposed \model with widely used and strong baselines:

\noindent \textit{\textbf{Reference-based Methods:}}

\noindent (1) \texttt{ROUGE}~\cite{Lin2004Rouge} is one of the most popular metrics, and it computes n-gram overlapping between the system output and reference summary. We employ the \texttt{ROUGE-1}, \texttt{ROUGE-2}, and \texttt{ROUGE-L} in our experiments.
(2) \texttt{BERTScore}~\cite{Zhang2020BERTScore} leverages the contextual embedding from the pre-training language model BERT~\cite{Devlin2019BERT} and calculates the cosine similarity between system output and reference.
(3) \texttt{MoverScore}~\cite{Zhao2019MoverScore} utilizes the Word Mover's Distance to compute the distance between the embedding of generated summary and reference.
(4) \texttt{BARTScore}~\cite{Yuan2021BARTScore} uses the weighted log probability of the pre-train language model BART's~\cite{lewis-etal-2020-bart} output to evaluate the quality of summaries.
(5) \texttt{CTC}~\cite{Degn2021CTC} is a general evaluation framework for language generation tasks including compression, transduction, and creation tasks. 
\texttt{CTC} is designed on the concept of information alignment. 
(6) \texttt{UniEval}~\cite{Zhong2022UniEval} formulates the summary evaluation as binary question answering and can evaluate the summary from four dimensions, coherence, consistency, fluency, and relevance. 

\noindent \textit{\textbf{Reference-free Methods:}}

\noindent (1) \texttt{BLANC}~\cite{Oleg2020BLANC} is defined as a measure of the helpfulness of a summary to PLM while PLM performs the Cloze task on document sentences. In specific, the final score is the accuracy difference of whether use a summary to concatenate with the masked sentence.
(2) \texttt{SummaQA}~\cite{Thomas2019SummaQA} is a QA-based evaluation metric. It generates questions from documents, answers the questions based on the summary by a QA model, and computes the QA metric as evaluation scores.
(3) \texttt{SUPERT}~\cite{Gao2020SUPERT} constructs the pseudo reference by extracting salient sentences from the source document and computes the similarity between generated summary and pseudo reference to evaluate the quality of the summary.
(4) \texttt{UniTE}~\cite{Yu2022UniTE} is a unified evaluation model for machine translation in different scenarios: reference-only, source-only and source-reference-combined.

To prove the effectiveness of the perturbed prefix-tuning, we design an ablation model, \model-PT (w/o \underline{P}refix-\underline{T}uning).
We remove the prefix of input and jointly fine-tune one pre-train language model using the two datasets we constructed.

\subsection{Implementation Details} 
Following \cite{Degn2021CTC}, we employ the roberta-large \cite{2019Roberta} as the backbone of our model. 
The MLP consists of 3 linear layers with tangent activation and the dimensions of each layer are 3072, 1024, and 2, respectively. 
Following \cite{Yu2022UniTE}, the max length of input sequence (with prompt) is set to 512.
We vary the length of prompt in \{8, 16, 32, 64, 128\}, and find that 128 is the best choice.
We use AdamW as the optimizer and the learning rate is set to 3.0e-05 selected from \{2.0e-05, 3.0e-05, 5.0e-05\}.
The number of train epochs is set up to 10 epochs and the batch size is set to 8. 
We fix the random seed always to 12 and trained our model on an NVIDIA GeForce RTX 3090 GPU for 6-7 hours.
We use PyLucene to implement the BM25 algorithm to retrieve similar documents.
The size of the two training datasets is 30K respectively, and the positive and negative samples are half.


\subsection{Evaluation Results} \label{sec:exp-result}


\begin{table*}[!ht]
\centering
\small
\resizebox{0.9\linewidth}{!}{
\begin{tabular}{ccccccccc}
\toprule
\multirow{2}{*}{\textbf{Model}} & \multicolumn{2}{c}{\textbf{Coherence}} & \multicolumn{2}{c}{\textbf{Consistency}} & \multicolumn{2}{c}{\textbf{Fluency}} & \multicolumn{2}{c}{\textbf{Relevance}} \\ 
\cmidrule(r){2-3}  \cmidrule(r){4-5} \cmidrule(r){6-7} \cmidrule(r){8-9}  
                       & $\rho$        & $\tau$        & $\rho$         & $\tau$         & $\rho$       & $\tau$       & $\rho$        & $\tau$        \\ \midrule
\textit{Sum-Ref Methods} \\

\texttt{ROUGE-1}~\cite{Lin2004Rouge}                  & 0.1670          & 0.1260          & 0.1600          & 0.1300          & 0.1590          & 0.0940          & 0.3260          & 0.2520          \\
\texttt{ROUGE-2}~\cite{Lin2004Rouge}                    & 0.1840          & 0.1390          & 0.1870          & 0.1550          & 0.1590          & 0.1280          & 0.2900          & 0.2190          \\
\texttt{ROUGE-L}~\cite{Lin2004Rouge}                   & 0.1280          & 0.0990          & 0.1150          & 0.0920          & 0.1050          & 0.0840          & 0.3110          & 0.2370          \\
\ding{228} \texttt{BERTScore}~\cite{Zhang2020BERTScore}                  & 0.2840          & 0.2110          & 0.1100          & 0.0900          & 0.1930          & 0.1580          & 0.3120          & 0.2430          \\
\texttt{MOVERScore}~\cite{Zhao2019MoverScore}                 & 0.1590          & 0.1180          & 0.1570          & 0.1270          & 0.1290          & 0.1050          & 0.3180          & 0.2440          \\
\texttt{UniTE (w/ SR)}\cite{Yu2022UniTE}	& 0.1792 	& 0.1362 	& 0.0557 	& 0.0474 	& 0.0761 	&0.0614 &	0.2255  & 0.1716 \\
\model (w/ SR)       & \underline{0.5840}\dubbelop          & \underline{0.4443}\dubbelop          & \underline{0.2494}\dubbelop          & \underline{0.2055}\dubbelop          & \underline{0.2601}\dubbelop          & \underline{0.2132}\dubbelop          & \underline{0.4217}\dubbelop          & \underline{0.3189}\dubbelop          \\
\sbbkgrnd \texttt{UniEval}~\cite{Zhong2022UniEval}         & \sbbkgrnd 0.4950 & \sbbkgrnd 0.3740 & \sbbkgrnd \textbf{0.4350} & \sbbkgrnd \textbf{0.3650} & \sbbkgrnd \textbf{0.4190} & \sbbkgrnd \textbf{0.3460} & \sbbkgrnd 0.4240 & \sbbkgrnd 0.3270  \\ 
\sbbkgrnd \texttt{BARTScore}~\cite{Yuan2021BARTScore}     & \sbbkgrnd 0.4480    & \sbbkgrnd 0.3420       & \sbbkgrnd 0.3820          & \sbbkgrnd 0.3150     & \sbbkgrnd 0.3560 & \sbbkgrnd 0.2920 & \sbbkgrnd 0.3560    & \sbbkgrnd 0.2730          \\ 
\midrule
\textit{Sum-Doc Methods}\\
\texttt{BLANC}~\cite{Oleg2020BLANC}                    & 0.1219          & 0.0951          & 0.2768          & 0.2307          & 0.1727          & 0.1436          & 0.2574          & 0.1983          \\
\texttt{SummaQA}~\cite{Thomas2019SummaQA}                   & 0.1239          & 0.0963          & 0.2540          & 0.2102          & 0.1782          & 0.1457          & 0.2120          & 0.1628          \\
\ding{228} \texttt{SUPERT}~\cite{Gao2020SUPERT}                    & 0.2165          & 0.1716          & 0.3438          & 0.2863          & 0.2509          & 0.2024          & 0.2746          & 0.2132          \\ 
\texttt{UniTE (w/ SD)}\cite{Yu2022UniTE}	& 0.1703 	& 0.1327 	& 0.1160 	& 0.0956 	& 0.0871 	& 0.0703 &	0.2738 	& 0.2084 \\

\model (w/ SD)           & \underline{0.5298}\dubbelop          & \underline{0.4052}\dubbelop          & \underline{0.3579}\dubbelop          & \underline{0.2961}\dubbelop         & \underline{0.3163}\dubbelop          & \underline{0.2617}\dubbelop          & \underline{0.4039}\dubbelop          & \underline{0.3060}\dubbelop          \\
\midrule
\textit{Sum-Doc-Ref Methods}\\
\ding{228} \texttt{CTC}~\cite{Degn2021CTC}                       & 0.4020          & 0.3100          & \underline{0.3660}          & \underline{0.3010}          & 0.2990          & 0.2450          & 0.4280          & 0.3360          \\
\texttt{UniTE (w/ SDR)}~\cite{Yu2022UniTE}                     & 0.1885          & 0.1453          & 0.1244          & 0.1017          & 0.1076          & 0.0886          &0.2874           & 0.2232  \\
\model (w/ SDR) & 0.4704 & 0.3532 & 0.3413 & 0.2817 & 0.3006 & 0.2451 & 0.3894 & 0.2929 \\
\model {\scriptsize (Fusion)} (w/ SDR) & \underline{\textbf{0.5944}}\dubbelop & \underline{\textbf{0.4515}}\dubbelop & 0.3381          & 0.2813          & \underline{0.3316}\dubbelop          & \underline{0.2731}\dubbelop          & \underline{\textbf{0.4358}}\dubbelop & \underline{\textbf{0.3282}}\dubbelop \\
\midrule
\textit{Ablation Methods}\\
\model-PT  (w/ SR)   & 0.5607 & 0.4246 & 0.2664 & 0.2193 & 0.2552 & 0.2079 &0.4228  & 0.3155      \\
\model-PT  (w/ SD)   &  0.5007 & 0.3810 & 0.3505 & 0.2905 & 0.3079 & 0.2533 & 0.4276 & 0.3220          \\
\model {\scriptsize (Fusion)}-PT (w/ SDR) & 0.5757 & 0.4397 & 0.3338 & 0.2751 & 0.3206 & 0.2638 & 0.4375 & 0.3291  \\
\bottomrule
\end{tabular}
}
\caption{Comparing with baselines on SummEval dataset. We use the notion ``(w/ *)'' to denote which data is used as input. $(\rho)$ denotes the Spearman correlations and $(\tau)$ denotes the Kendall-Tau correlations. The row with \colorbox{gray!35}{shaded background} denotes the multi-dimensional metrics which output a score for each dimension, and it is unfair for comparing with these methods. The number with \underline{underline} denotes the max value in the scenario and the \textbf{bold-face} denotes the max value over three scenarios.}
\label{tab:overall-exp-result}
\end{table*}

\begin{table*}[!ht]
\centering
\small
\resizebox{1.5\columnwidth}{!}{
\begin{tabular}{ccccccccc}
\toprule
 \multirow{2}{*}{\textbf{Fusion Methods}} & \multicolumn{2}{c}{\textbf{Coherence}} & \multicolumn{2}{c}{\textbf{Consistency}} & \multicolumn{2}{c}{\textbf{Fluency}} & \multicolumn{2}{c}{\textbf{Relevance}} \\
\cmidrule(r){2-3}  \cmidrule(r){4-5} \cmidrule(r){6-7} \cmidrule(r){8-9}  
 & $\rho$             & $\tau$            & $\rho$               & $\tau$            & $\rho$            & $\tau$           & $\rho$             & $\tau$            \\  \midrule
Min & 0.5896	& 0.4475	& 0.2729	& 0.2261	& 0.2786	& 0.2284	& 0.4315	& 0.3267 \\
Max & 0.5386	& 0.4099	& \textbf{0.3561}	& \textbf{0.2947}	& 0.3183	& 0.2624	& 0.4083	& 0.3084 \\
Geometric mean & 0.5938	& 0.4503	& 0.3151 & 0.2618	& 0.3132	& 0.2584	& 0.4332 & 	0.3260 \\ 
Arithmetic mean \ding{51} & \textbf{0.5944}	& \textbf{0.4515}	& 0.3381	& 0.2813	& \textbf{0.3316}	& \textbf{0.2731}	& \textbf{0.4358}  & \textbf{0.3282} \\
\bottomrule
\end{tabular}
}
\caption{Result of different fusion methods in Sum-Doc-Ref scenario.}
\label{tab:exp-fusion-variant}
\end{table*}

\newcommand{\dd}{$\downarrow$}
\newcommand{\uu}{$\uparrow$}

\begin{table*}[!ht]
\centering
\small
\resizebox{1.8\columnwidth}{!}{
\begin{tabular}{ccccccccc}
\toprule
 \multirow{2}{*}{} & \multicolumn{2}{c}{\textbf{Coherence}} & \multicolumn{2}{c}{\textbf{Consistency}} & \multicolumn{2}{c}{\textbf{Fluency}} & \multicolumn{2}{c}{\textbf{Relevance}} \\
\cmidrule(r){2-3}  \cmidrule(r){4-5} \cmidrule(r){6-7} \cmidrule(r){8-9}  
 & $\rho$             & $\tau$            & $\rho$               & $\tau$            & $\rho$            & $\tau$           & $\rho$             & $\tau$            \\  
 \midrule
Single Model (w/ SR)       & 0.5019          & 0.3796          & \underline{0.2916}           & \underline{0.2391}          & \underline{0.3090}           & \underline{0.2525}          & 0.4153           & 0.3096 \\
\model (w/ SR)       & \underline{0.5840}\uu          & \underline{0.4443}\uu          & 0.2494\dd          & 0.2055\dd          & 0.2601\dd          & 0.2132\dd          & \underline{0.4217}\uu          & \underline{0.3189}\uu          \\
\midrule
Single Model (w/ SD)       &0.4798         & 0.3599         & 0.3132         & 0.2580         & 0.2992         & 0.2454         & 0.3644         & 0.2760 \\
\model (w/ SD)              & \underline{0.5298}\uu          & \underline{0.4052}\uu          & \underline{0.3579}\uu          & \underline{0.2961}\uu         & \underline{0.3163}\uu          & \underline{0.2617}\uu          & \underline{0.4039}\uu          & \underline{0.3060}\uu          \\
\midrule

Single Model (w/ SDR)              & 0.3488 	& 0.2660 	& 0.2824 	& 0.2342 	& 0.2739 &	0.2253 & 	0.2435 	& 0.1866 \\
\model (w/ SDR) & \underline{0.4704}\uu & \underline{0.3532}\uu & \underline{0.3413}\uu & \underline{0.2817}\uu & \underline{0.3006}\uu & \underline{0.2451}\uu & \underline{0.3894}\uu & \underline{0.2929}\uu \\

\bottomrule
\end{tabular}
}
\vspace{-1mm}
\caption{Comparison between \model and separately fine-tuning PLM.}
\vspace{-4mm}
\label{tab:exp-comp-single}
\end{table*}

We compare our \model with strong baselines in Table~\ref{tab:overall-exp-result}.
We can surprisingly find that \model (w/ SD) performs comparably to the \model (w/ SR) in the Sum-Ref scenario and achieves significant improvement over the existing baselines, which demonstrates that our proposed perturbed prefix-tuning can transfer knowledge from other scenarios.
\texttt{BERTScore} is the state-of-the-art reference-based single-dimensional evaluation method, and the performance of \model increases by 105.63\%, 34.93\%, and 38.62\% compared to \texttt{BERTScore} in terms of Coherence ($\rho$), Fluency ($\tau$), and Relevance ($\rho$) respectively.
Compared with the reference-free baselines, \model (w/ SD) outperforms \texttt{SUPERT} 144.71\%, 29.30\%, and 47.09\%  in terms of Coherence ($\rho$), Fluency ($\tau$), and Relevance ($\rho$) respectively.
Although the \model achieves slightly lower performance than the baseline in one dimension, the \model achieves consistently strong performance in three scenarios which can facilitate users from having to use multiple models.

As illustrated in the related work \S~\ref{sec:related}, some evaluators (\eg \texttt{UniEval} and \texttt{BARTScore}) focus on evaluating the summary in multi-dimension which model the specific dimension features and output \textit{multiple scores}.
Different from these methods, we focus on an orthogonal aspect that uses a unified model in multiple scenarios, and we only use \textit{one score} to represent the summary quality.
Thus, directly comparing with these multi-dimensional metrics is not fair.
Since our unified multi-scenario evaluator is orthogonal to these multi-dimension evaluators, we will combine the multi-dimensional method into \model in future work.

Similar to our \model, \texttt{UniTE} is also a multi-scenario unified evaluation method for machine translation.
However, \texttt{UniTE} achieves worse performance than \model, which demonstrates our assumption that the matching framework and the data construction method in \texttt{UniTE} are mainly focusing on the characteristic of translation. 
And we cannot simply use \texttt{UniTE} in the summarization task. 

From the results of \model {\scriptsize (Fusion)} (w/ SDR) and \model (w/ SDR), we can find that the fusion model achieves better performance, and we will use the fusion method in our release version of \model.
An extensive analysis of why the fusion method works better than directly concatenating Sum-Doc-Ref in the input of PLM is shown in the following section.

\begin{table}[!t]
\centering
\small
\begin{tabular}{ccc}
\toprule
\textbf{Model} & \textbf{Faithful} & \textbf{Factual} \\
\midrule
ROUGE-1    & 0.197    & 0.125     \\
ROUGE-2    & 0.162    & 0.095     \\
ROUGE-L    & 0.162    & 0.113     \\
BERTScore  & 0.190    & 0.116     \\
QA         & 0.044	  & 0.027     \\
UMSE	   & \textbf{0.242}	  & \textbf{0.167}     \\
\bottomrule
\sbbkgrnd  Entailment & \sbbkgrnd  0.431	  &  \sbbkgrnd 0.264     \\
\toprule
\end{tabular}
\caption{
The performance of different models on detecting hallucinations. The evaluation metric is the Spearman correlation. The faithful and factual annotations are released by~\citet{Joshua2022Faith_Fact_Abs_Summ}.
The row with \colorbox{gray!35}{shaded background} denotes the model is trained on a supervised dataset, making it unfair to compare it with other methods.}
\label{tab:hallucination}
\end{table}

\subsection{Discussions}
\paragraph{Ablation Studies}\label{sec:ablation-exp}
To verify the effectiveness of our proposed perturbed prefix tuning method, we employ an ablation model \model-PT in three scenarios.
In this model, we mix the training datasets we constructed and jointly fine-tune one PLM for 
\textit{all} scenarios.
From the results shown in Table~\ref{tab:overall-exp-result}, we can find that \model-PT underperforms with the \model in all scenarios.
Although using a shared pre-train language model can also transfer knowledge among these scenarios, these ablation studies demonstrate that using the shared continuous prefix tokens provides an explicit way to share common matching knowledge and it can boost the performance of the \model.

Moreover, we employ an intuitive experiment that separately fine-tunes a PLM for \textit{each} scenario, and the results are shown in Table~\ref{tab:exp-comp-single}.
Although the performance of the Sum-Ref drops slightly in terms of two dimensions, our proposed \model boosts the performance in the Sum-Doc scenario significantly.
And boosting the performance of the Sum-Doc scenario is more valuable since evaluation in this scenario does not require any human annotating.

\vspace{-1mm}
\paragraph{Analysis of Sum-Doc-Ref Fusion}
In \S~\ref{sec:sdr-variant}, we propose a variant model for the Sum-Doc-Ref scenario which directly fuses the scores of Sum-Doc and Sum-Ref to produce the score for the Sum-Doc-Ref scenario.
In this section, we conduct experiments to explore which fusion method will lead to better performance.
We employ four different fusion methods:
(1) max method takes the maximum of $s_{SD}$ and $s_{SR}$ as $s_{SDR}$;
(2) min method takes the minimum of $s_{SD}$ and $s_{SR}$;
(3) geometric mean fusion uses $\sqrt{s_{SD} s_{SR}}$ as $s_{SDR}$;
and (4) arithmetic mean fusion employs $\textstyle \frac{(s_{SD}+s_{SR})}{2}$.
From Table~\ref{tab:exp-fusion-variant}, we can find that the arithmetic mean achieves the best performance, and we finally use the arithmetic mean fusion in the \model {\scriptsize (Fusion)}.

\begin{figure}
\centering
\includegraphics[width=0.99\columnwidth]{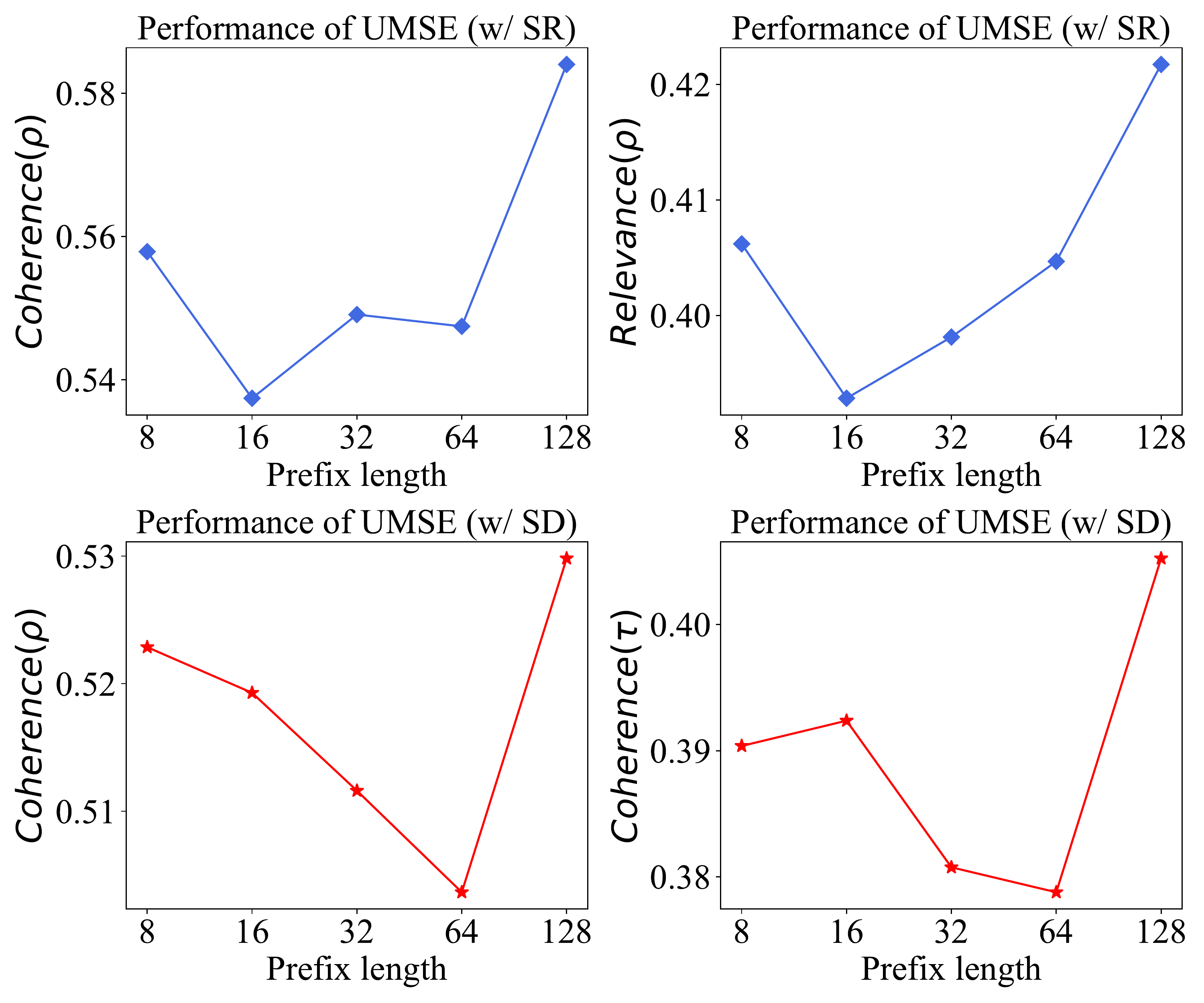}
\vspace{-6mm}
\caption{Performance across different prefix lengths.}
\vspace{-5mm}
\label{fig:diff-len-prefix}
\end{figure}

\vspace{-1mm}
\paragraph{Analysis of Perturbed Prefix Length}
To verify the effectiveness of our proposed perturbed prefix, we conduct experiments using the different lengths of the prefix.
From Figure~\ref{fig:diff-len-prefix}, we can find that the performance of our \model gradually improved with the growth of the prefix length.
\paragraph{Analysis of Hallucination Detection}
To analyze the effectiveness of our model in detecting hallucinations, we conducted experiments on the dataset released by~\citet{Joshua2022Faith_Fact_Abs_Summ}
and the results are shown in Table~\ref{tab:hallucination}. 
According to the Spearman correlations on both faithful and factual, UMSE outperforms baselines, such as ROUGE, BERTScore, and QA, which demonstrates the ability of our proposed model in detecting hallucinations.

\section{Conclusion}

In this paper, we propose \fullmodel (\model) which is a unified multi-scenario summarization evaluation framework.
\model can perform the semantic evaluation on three typical evaluation scenarios: (1) Sum-Ref; (2) Sum-Doc and (3) Sum-Doc-Ref using only one unified model.
Since these scenarios have different input formats, we propose a perturbed prefix-tuning method that unifies these different scenarios in one model and it can also transfer knowledge between these scenarios.
To train the \model in a self-supervised manner, we propose two training data construction methods without using any human annotation.
Extensive experiments conducted on the benchmark dataset SummEval verify that the \model can achieve comparable performance with existing baselines.

\section*{Limitations}
In this paper, we propose the evaluation model \model which can be used to evaluate the summary quality in three typical scenarios.
However, 
in the summarization task, different annotators have different writing styles, and there might exist more than one good summary for one document.
Moreover, there can be summaries that concentrate on different aspects of a document (\eg describing the location and room of a hotel).
In the future, we aim to incorporate more scenarios (\eg multi-references and multi-aspects) into our unified evaluation method.

\section*{Ethics Statement}
In this section, we would like to discuss the ethical concerns of our work. Our proposed method \model is a unified model for multi-scenario summarization evaluation and is designed to help humans efficiently evaluate summaries.
And the sensitive information is masked while constructing the training data from CNN/DailyMail dataset.

\section*{Acknowledgements}
We would like to express sincere thanks to the anonymous reviewers for their helpful comments. This research was supported by the Natural Science Foundation of China (T2293773, 62102234, 62272274, 62202271, 61902219, 61972234, 62072279), the National Key R\&D Program of China with grant (No.2022YFC3303004, No.2020YFB1406704), the Key Scientific and Technological Innovation Program of Shandong Province (2019JZZY010129), 
the Tencent WeChat Rhino-Bird Focused Research Program (JR-WXG-2021411), 
the Fundamental Research Funds of Shandong University.

\bibliography{anthology,references}

\begin{thebibliography}{34}
\expandafter\ifx\csname natexlab\endcsname\relax\def\natexlab#1{#1}\fi

\bibitem[{Banerjee and Lavie(2005)}]{Banerjee2005METEOR}
Satanjeev Banerjee and Alon Lavie. 2005.
\newblock \href {https://aclanthology.org/W05-0909} {{METEOR}: An automatic
  metric for {MT} evaluation with improved correlation with human judgments}.
\newblock In \emph{Proceedings of the {ACL} Workshop on Intrinsic and Extrinsic
  Evaluation Measures for Machine Translation and/or Summarization}, pages
  65--72, Ann Arbor, Michigan. Association for Computational Linguistics.

\bibitem[{Chen et~al.(2021)Chen, Li, and King}]{Chen2021Multi_Doc_Metric}
Wang Chen, Piji Li, and Irwin King. 2021.
\newblock \href {https://doi.org/10.18653/v1/2021.acl-long.34} {A training-free
  and reference-free summarization evaluation metric via centrality-weighted
  relevance and self-referenced redundancy}.
\newblock In \emph{Proceedings of the 59th Annual Meeting of the Association
  for Computational Linguistics and the 11th International Joint Conference on
  Natural Language Processing (Volume 1: Long Papers)}, pages 404--414, Online.
  Association for Computational Linguistics.

\bibitem[{Deng et~al.(2021)Deng, Tan, Liu, Xing, and Hu}]{Degn2021CTC}
Mingkai Deng, Bowen Tan, Zhengzhong Liu, Eric Xing, and Zhiting Hu. 2021.
\newblock \href {https://doi.org/10.18653/v1/2021.emnlp-main.599} {Compression,
  transduction, and creation: A unified framework for evaluating natural
  language generation}.
\newblock In \emph{Proceedings of the 2021 Conference on Empirical Methods in
  Natural Language Processing}, pages 7580--7605, Online and Punta Cana,
  Dominican Republic. Association for Computational Linguistics.

\bibitem[{Devlin et~al.(2019)Devlin, Chang, Lee, and
  Toutanova}]{Devlin2019BERT}
Jacob Devlin, Ming-Wei Chang, Kenton Lee, and Kristina Toutanova. 2019.
\newblock \href {https://doi.org/10.18653/v1/N19-1423} {{BERT}: Pre-training of
  deep bidirectional transformers for language understanding}.
\newblock In \emph{Proceedings of the 2019 Conference of the North {A}merican
  Chapter of the Association for Computational Linguistics: Human Language
  Technologies, Volume 1 (Long and Short Papers)}, pages 4171--4186,
  Minneapolis, Minnesota. Association for Computational Linguistics.

\bibitem[{Durmus et~al.(2020)Durmus, He, and Diab}]{Esin2020FEQA}
Esin Durmus, He~He, and Mona Diab. 2020.
\newblock \href {https://doi.org/10.18653/v1/2020.acl-main.454} {{FEQA}: A
  question answering evaluation framework for faithfulness assessment in
  abstractive summarization}.
\newblock In \emph{Proceedings of the 58th Annual Meeting of the Association
  for Computational Linguistics}, pages 5055--5070, Online. Association for
  Computational Linguistics.

\bibitem[{Fabbri et~al.(2021)Fabbri, Kry{\'s}ci{\'n}ski, McCann, Xiong, Socher,
  and Radev}]{Fabbri2021Summeval}
Alexander~R. Fabbri, Wojciech Kry{\'s}ci{\'n}ski, Bryan McCann, Caiming Xiong,
  Richard Socher, and Dragomir Radev. 2021.
\newblock \href {https://doi.org/10.1162/tacl_a_00373} {{S}umm{E}val:
  Re-evaluating summarization evaluation}.
\newblock \emph{Transactions of the Association for Computational Linguistics},
  9:391--409.

\bibitem[{Gao et~al.(2019{\natexlab{a}})Gao, Chen, Li, Ren, Bing, Zhao, and
  Yan}]{Gao2019RASG}
Shen Gao, Xiuying Chen, Piji Li, Zhaochun Ren, Lidong Bing, Dongyan Zhao, and
  Rui Yan. 2019{\natexlab{a}}.
\newblock \href {https://doi.org/10.1609/aaai.v33i01.33016399} {Abstractive
  text summarization by incorporating reader comments}.
\newblock In \emph{The Thirty-Third {AAAI} Conference on Artificial
  Intelligence, {AAAI} 2019, The Thirty-First Innovative Applications of
  Artificial Intelligence Conference, {IAAI} 2019, The Ninth {AAAI} Symposium
  on Educational Advances in Artificial Intelligence, {EAAI} 2019, Honolulu,
  Hawaii, USA, January 27 - February 1, 2019}, pages 6399--6406. {AAAI} Press.

\bibitem[{Gao et~al.(2020{\natexlab{a}})Gao, Chen, Ren, Zhao, and
  Yan}]{Gao2020Summarization}
Shen Gao, Xiuying Chen, Zhaochun Ren, Dongyan Zhao, and Rui Yan.
  2020{\natexlab{a}}.
\newblock \href {https://doi.org/10.24963/ijcai.2020/676} {From standard
  summarization to new tasks and beyond: Summarization with manifold
  information}.
\newblock In \emph{Proceedings of the Twenty-Ninth International Joint
  Conference on Artificial Intelligence, {IJCAI} 2020}, pages 4854--4860.
  ijcai.org.

\bibitem[{Gao et~al.(2019{\natexlab{b}})Gao, Ren, Zhao, Zhao, Yin, and
  Yan}]{Gao2019PAAG}
Shen Gao, Zhaochun Ren, Yihong~Eric Zhao, Dongyan Zhao, Dawei Yin, and Rui Yan.
  2019{\natexlab{b}}.
\newblock \href {https://doi.org/10.1145/3289600.3290992} {Product-aware answer
  generation in e-commerce question-answering}.
\newblock In \emph{Proceedings of the Twelfth {ACM} International Conference on
  Web Search and Data Mining, {WSDM} 2019, Melbourne, VIC, Australia, February
  11-15, 2019}, pages 429--437. {ACM}.

\bibitem[{Gao et~al.(2020{\natexlab{b}})Gao, Zhao, and Eger}]{Gao2020SUPERT}
Yang Gao, Wei Zhao, and Steffen Eger. 2020{\natexlab{b}}.
\newblock \href {https://doi.org/10.18653/v1/2020.acl-main.124} {{SUPERT}:
  Towards new frontiers in unsupervised evaluation metrics for multi-document
  summarization}.
\newblock In \emph{Proceedings of the 58th Annual Meeting of the Association
  for Computational Linguistics}, pages 1347--1354, Online. Association for
  Computational Linguistics.

\bibitem[{Lewis et~al.(2020)Lewis, Liu, Goyal, Ghazvininejad, Mohamed, Levy,
  Stoyanov, and Zettlemoyer}]{lewis-etal-2020-bart}
Mike Lewis, Yinhan Liu, Naman Goyal, Marjan Ghazvininejad, Abdelrahman Mohamed,
  Omer Levy, Veselin Stoyanov, and Luke Zettlemoyer. 2020.
\newblock \href {https://doi.org/10.18653/v1/2020.acl-main.703} {{BART}:
  Denoising sequence-to-sequence pre-training for natural language generation,
  translation, and comprehension}.
\newblock In \emph{Proceedings of the 58th Annual Meeting of the Association
  for Computational Linguistics}, pages 7871--7880, Online. Association for
  Computational Linguistics.

\bibitem[{Li and Liang(2021)}]{Liang2021Prefix}
Xiang~Lisa Li and Percy Liang. 2021.
\newblock \href {https://doi.org/10.18653/v1/2021.acl-long.353} {Prefix-tuning:
  Optimizing continuous prompts for generation}.
\newblock In \emph{Proceedings of the 59th Annual Meeting of the Association
  for Computational Linguistics and the 11th International Joint Conference on
  Natural Language Processing (Volume 1: Long Papers)}, pages 4582--4597,
  Online. Association for Computational Linguistics.

\bibitem[{Lin(2004)}]{Lin2004Rouge}
Chin-Yew Lin. 2004.
\newblock \href {https://aclanthology.org/W04-1013} {{ROUGE}: A package for
  automatic evaluation of summaries}.
\newblock In \emph{Text Summarization Branches Out}, pages 74--81, Barcelona,
  Spain. Association for Computational Linguistics.

\bibitem[{Liu et~al.(2021)Liu, Ji, Fu, Du, Yang, and Tang}]{Liu2021Ptuningv2}
Xiao Liu, Kaixuan Ji, Yicheng Fu, Zhengxiao Du, Zhilin Yang, and Jie Tang.
  2021.
\newblock P-tuning v2: Prompt tuning can be comparable to fine-tuning
  universally across scales and tasks.
\newblock \emph{CoRR}, abs/2110.07602.

\bibitem[{Liu et~al.(2022)Liu, Ji, Fu, Tam, Du, Yang, and
  Tang}]{Liu2022Ptuningv1}
Xiao Liu, Kaixuan Ji, Yicheng Fu, Weng Tam, Zhengxiao Du, Zhilin Yang, and Jie
  Tang. 2022.
\newblock \href {https://doi.org/10.18653/v1/2022.acl-short.8} {{P}-tuning:
  Prompt tuning can be comparable to fine-tuning across scales and tasks}.
\newblock In \emph{Proceedings of the 60th Annual Meeting of the Association
  for Computational Linguistics (Volume 2: Short Papers)}, pages 61--68,
  Dublin, Ireland. Association for Computational Linguistics.

\bibitem[{Liu et~al.(2019)Liu, Ott, Goyal, Du, Joshi, Chen, Levy, Lewis,
  Zettlemoyer, and Stoyanov}]{2019Roberta}
Yinhan Liu, Myle Ott, Naman Goyal, Jingfei Du, Mandar Joshi, Danqi Chen, Omer
  Levy, Mike Lewis, Luke Zettlemoyer, and Veselin Stoyanov. 2019.
\newblock \href {https://arxiv.org/abs/1907.11692} {Roberta: {A} robustly
  optimized {BERT} pretraining approach}.
\newblock \emph{ArXiv preprint}, abs/1907.11692.

\bibitem[{Maynez et~al.(2020)Maynez, Narayan, Bohnet, and
  McDonald}]{Joshua2022Faith_Fact_Abs_Summ}
Joshua Maynez, Shashi Narayan, Bernd Bohnet, and Ryan McDonald. 2020.
\newblock \href {https://doi.org/10.18653/v1/2020.acl-main.173} {On
  faithfulness and factuality in abstractive summarization}.
\newblock In \emph{Proceedings of the 58th Annual Meeting of the Association
  for Computational Linguistics}, pages 1906--1919, Online. Association for
  Computational Linguistics.

\bibitem[{Mikolov et~al.(2013)Mikolov, Sutskever, Chen, Corrado, and
  Dean}]{Tomas2013Word2vec}
Tom{\'{a}}s Mikolov, Ilya Sutskever, Kai Chen, Gregory~S. Corrado, and Jeffrey
  Dean. 2013.
\newblock \href
  {https://proceedings.neurips.cc/paper/2013/hash/9aa42b31882ec039965f3c4923ce901b-Abstract.html}
  {Distributed representations of words and phrases and their
  compositionality}.
\newblock In \emph{Advances in Neural Information Processing Systems 26: 27th
  Annual Conference on Neural Information Processing Systems 2013. Proceedings
  of a meeting held December 5-8, 2013, Lake Tahoe, Nevada, United States},
  pages 3111--3119.

\bibitem[{Nallapati et~al.(2016)Nallapati, Zhou, Gulcehre, Xiang
  et~al.}]{Nallapati2016CNN/DM}
Ramesh Nallapati, Bowen Zhou, Caglar Gulcehre, Bing Xiang, et~al. 2016.
\newblock \href {https://doi.org/10.18653/v1/K16-1028} {Abstractive text
  summarization using sequence-to-sequence {RNN}s and beyond}.
\newblock In \emph{Proceedings of the 20th {SIGNLL} Conference on Computational
  Natural Language Learning}, pages 280--290, Berlin, Germany. Association for
  Computational Linguistics.

\bibitem[{Narayan et~al.(2018)Narayan, Cohen, and Lapata}]{Sha2018RLforsum}
Shashi Narayan, Shay~B. Cohen, and Mirella Lapata. 2018.
\newblock \href {https://doi.org/10.18653/v1/N18-1158} {Ranking sentences for
  extractive summarization with reinforcement learning}.
\newblock In \emph{Proceedings of the 2018 Conference of the North {A}merican
  Chapter of the Association for Computational Linguistics: Human Language
  Technologies, Volume 1 (Long Papers)}, pages 1747--1759, New Orleans,
  Louisiana. Association for Computational Linguistics.

\bibitem[{Ng and Abrecht(2015)}]{NG2015Rougewe}
Jun-Ping Ng and Viktoria Abrecht. 2015.
\newblock \href {https://doi.org/10.18653/v1/D15-1222} {Better summarization
  evaluation with word embeddings for {ROUGE}}.
\newblock In \emph{Proceedings of the 2015 Conference on Empirical Methods in
  Natural Language Processing}, pages 1925--1930, Lisbon, Portugal. Association
  for Computational Linguistics.

\bibitem[{Papineni et~al.(2002)Papineni, Roukos, Ward, and
  Zhu}]{Papineni2002Bleu}
Kishore Papineni, Salim Roukos, Todd Ward, and Wei-Jing Zhu. 2002.
\newblock \href {https://doi.org/10.3115/1073083.1073135} {{B}leu: a method for
  automatic evaluation of machine translation}.
\newblock In \emph{Proceedings of the 40th Annual Meeting of the Association
  for Computational Linguistics}, pages 311--318, Philadelphia, Pennsylvania,
  USA. Association for Computational Linguistics.

\bibitem[{Peyrard et~al.(2017)Peyrard, Botschen, and Gurevych}]{Maxime2017S3}
Maxime Peyrard, Teresa Botschen, and Iryna Gurevych. 2017.
\newblock \href {https://doi.org/10.18653/v1/W17-4510} {Learning to score
  system summaries for better content selection evaluation.}
\newblock In \emph{Proceedings of the Workshop on New Frontiers in
  Summarization}, pages 74--84, Copenhagen, Denmark. Association for
  Computational Linguistics.

\bibitem[{Raffel et~al.(2020)Raffel, Shazeer, Roberts, Lee, Narang, Matena,
  Zhou, Li, and Liu}]{2020t5}
Colin Raffel, Noam Shazeer, Adam Roberts, Katherine Lee, Sharan Narang, Michael
  Matena, Yanqi Zhou, Wei Li, and Peter~J. Liu. 2020.
\newblock \href {http://jmlr.org/papers/v21/20-074.html} {Exploring the limits
  of transfer learning with a unified text-to-text transformer}.
\newblock \emph{J. Mach. Learn. Res.}, 21:140:1--140:67.

\bibitem[{Scialom et~al.(2021)Scialom, Dray, Lamprier, Piwowarski, Staiano,
  Wang, and Gallinari}]{Thomas2021QuestEval}
Thomas Scialom, Paul-Alexis Dray, Sylvain Lamprier, Benjamin Piwowarski, Jacopo
  Staiano, Alex Wang, and Patrick Gallinari. 2021.
\newblock \href {https://doi.org/10.18653/v1/2021.emnlp-main.529}
  {{Q}uest{E}val: Summarization asks for fact-based evaluation}.
\newblock In \emph{Proceedings of the 2021 Conference on Empirical Methods in
  Natural Language Processing}, pages 6594--6604, Online and Punta Cana,
  Dominican Republic. Association for Computational Linguistics.

\bibitem[{Scialom et~al.(2019)Scialom, Lamprier, Piwowarski, and
  Staiano}]{Thomas2019SummaQA}
Thomas Scialom, Sylvain Lamprier, Benjamin Piwowarski, and Jacopo Staiano.
  2019.
\newblock \href {https://doi.org/10.18653/v1/D19-1320} {Answers unite!
  unsupervised metrics for reinforced summarization models}.
\newblock In \emph{Proceedings of the 2019 Conference on Empirical Methods in
  Natural Language Processing and the 9th International Joint Conference on
  Natural Language Processing (EMNLP-IJCNLP)}, pages 3246--3256, Hong Kong,
  China. Association for Computational Linguistics.

\bibitem[{Vasilyev et~al.(2020)Vasilyev, Dharnidharka, and
  Bohannon}]{Oleg2020BLANC}
Oleg Vasilyev, Vedant Dharnidharka, and John Bohannon. 2020.
\newblock \href {https://doi.org/10.18653/v1/2020.eval4nlp-1.2} {Fill in the
  {BLANC}: Human-free quality estimation of document summaries}.
\newblock In \emph{Proceedings of the First Workshop on Evaluation and
  Comparison of NLP Systems}, pages 11--20, Online. Association for
  Computational Linguistics.

\bibitem[{Wan et~al.(2022)Wan, Liu, Yang, Zhang, Chen, Wong, and
  Chao}]{Yu2022UniTE}
Yu~Wan, Dayiheng Liu, Baosong Yang, Haibo Zhang, Boxing Chen, Derek Wong, and
  Lidia Chao. 2022.
\newblock \href {https://doi.org/10.18653/v1/2022.acl-long.558} {{U}ni{TE}:
  Unified translation evaluation}.
\newblock In \emph{Proceedings of the 60th Annual Meeting of the Association
  for Computational Linguistics (Volume 1: Long Papers)}, pages 8117--8127,
  Dublin, Ireland. Association for Computational Linguistics.

\bibitem[{Wu et~al.(2020)Wu, Ma, Wu, Manyumwa, and Ji}]{Wu2020LS_Score}
Hanlu Wu, Tengfei Ma, Lingfei Wu, Tariro Manyumwa, and Shouling Ji. 2020.
\newblock \href {https://doi.org/10.18653/v1/2020.emnlp-main.294} {Unsupervised
  reference-free summary quality evaluation via contrastive learning}.
\newblock In \emph{Proceedings of the 2020 Conference on Empirical Methods in
  Natural Language Processing (EMNLP)}, pages 3612--3621, Online. Association
  for Computational Linguistics.

\bibitem[{Wu and Hu(2018)}]{Wu2018RLforsum}
Yuxiang Wu and Baotian Hu. 2018.
\newblock \href
  {https://www.aaai.org/ocs/index.php/AAAI/AAAI18/paper/view/16838} {Learning
  to extract coherent summary via deep reinforcement learning}.
\newblock In \emph{Proceedings of the Thirty-Second {AAAI} Conference on
  Artificial Intelligence, (AAAI-18), the 30th innovative Applications of
  Artificial Intelligence (IAAI-18), and the 8th {AAAI} Symposium on
  Educational Advances in Artificial Intelligence (EAAI-18), New Orleans,
  Louisiana, USA, February 2-7, 2018}, pages 5602--5609. {AAAI} Press.

\bibitem[{Yuan et~al.(2021)Yuan, Neubig, and Liu}]{Yuan2021BARTScore}
Weizhe Yuan, Graham Neubig, and Pengfei Liu. 2021.
\newblock \href
  {https://proceedings.neurips.cc/paper/2021/hash/e4d2b6e6fdeca3e60e0f1a62fee3d9dd-Abstract.html}
  {Bartscore: Evaluating generated text as text generation}.
\newblock In \emph{Advances in Neural Information Processing Systems 34: Annual
  Conference on Neural Information Processing Systems 2021, NeurIPS 2021,
  December 6-14, 2021, virtual}, pages 27263--27277.

\bibitem[{Zhang et~al.(2020)Zhang, Kishore, Wu, Weinberger, and
  Artzi}]{Zhang2020BERTScore}
Tianyi Zhang, Varsha Kishore, Felix Wu, Kilian~Q. Weinberger, and Yoav Artzi.
  2020.
\newblock \href {https://openreview.net/forum?id=SkeHuCVFDr} {Bertscore:
  Evaluating text generation with {BERT}}.
\newblock In \emph{8th International Conference on Learning Representations,
  {ICLR} 2020, Addis Ababa, Ethiopia, April 26-30, 2020}. OpenReview.net.

\bibitem[{Zhao et~al.(2019)Zhao, Peyrard, Liu, Gao, Meyer, and
  Eger}]{Zhao2019MoverScore}
Wei Zhao, Maxime Peyrard, Fei Liu, Yang Gao, Christian~M. Meyer, and Steffen
  Eger. 2019.
\newblock \href {https://doi.org/10.18653/v1/D19-1053} {{M}over{S}core: Text
  generation evaluating with contextualized embeddings and earth mover
  distance}.
\newblock In \emph{Proceedings of the 2019 Conference on Empirical Methods in
  Natural Language Processing and the 9th International Joint Conference on
  Natural Language Processing (EMNLP-IJCNLP)}, pages 563--578, Hong Kong,
  China. Association for Computational Linguistics.

\bibitem[{Zhong et~al.(2022)Zhong, Liu, Yin, Mao, Jiao, Liu, Zhu, Ji, and
  Han}]{Zhong2022UniEval}
Ming Zhong, Yang Liu, Da~Yin, Yuning Mao, Yizhu Jiao, Pengfei Liu, Chenguang
  Zhu, Heng Ji, and Jiawei Han. 2022.
\newblock \href {https://aclanthology.org/2022.emnlp-main.131} {Towards a
  unified multi-dimensional evaluator for text generation}.
\newblock In \emph{Proceedings of the 2022 Conference on Empirical Methods in
  Natural Language Processing}, pages 2023--2038, Abu Dhabi, United Arab
  Emirates. Association for Computational Linguistics.

\end{thebibliography}
\bibliographystyle{acl_natbib}




\end{document}